\pdfoutput=1

\documentclass[11pt]{article}

\usepackage{EMNLP2023}
 
\usepackage{times}
\usepackage{latexsym}

\usepackage[T1]{fontenc}

\usepackage[utf8]{inputenc}

\usepackage{microtype}

\usepackage{inconsolata}
\usepackage[ruled,vlined]{algorithm2e}

\usepackage{xspace}
\newcommand{\methodname}[1]{{\textsc{Fabricator}}}

\usepackage{graphicx}
\usepackage{booktabs}
\usepackage{multirow}
\usepackage{cleveref}
\usepackage{listings}
\usepackage{xcolor}
\usepackage{enumitem}

\definecolor{codegreen}{rgb}{0,0.6,0}
\definecolor{codegray}{rgb}{0.5,0.5,0.5}
\definecolor{codepurple}{rgb}{0.58,0,0.82}
\definecolor{backcolour}{rgb}{0.95,0.95,0.92}
\lstdefinestyle{mystyle}{
    backgroundcolor=\color{backcolour},   
    commentstyle=\color{codegreen},
    keywordstyle=\color{magenta},
    numberstyle=\tiny\color{codegray},
    stringstyle=\color{codepurple},
    basicstyle=\ttfamily\footnotesize,
    breakatwhitespace=false,         
    breaklines=true,                 
    captionpos=b,                    
    keepspaces=true,                 
    numbers=left,                    
    numbersep=5pt,                  
    showspaces=false,                
    showstringspaces=false,
    showtabs=false,                  
    tabsize=2,
    xleftmargin=1.8em,
}

\title{\methodname{}: An Open Source Toolkit for Generating\\ Labeled Training Data with Teacher LLMs}

\author{Jonas Golde$^1$, Patrick Haller$^1$, Felix Hamborg$^1$, Julian Risch$^2$, Alan Akbik$^1$ \\
  $^1$ Humboldt University of Berlin \\
  $^2$ deepset GmbH \\
  \texttt{\{jonas.golde, patrick.haller.1, felix.hamborg, alan.akbik\}@hu-berlin.de}\\
  \texttt{julian.risch@deepset.ai}
}

\begin{document}
{\makeatletter\acl@finalcopytrue
  \maketitle
}
\begin{abstract}
Most NLP tasks are modeled as supervised learning and thus require labeled training data to train effective models. 
However, manually producing such data at sufficient quality and quantity is known to be costly and time-intensive. 
Current research addresses this bottleneck by exploring a novel paradigm called \textit{zero-shot learning via dataset generation}. Here, a powerful LLM is prompted with a task description to generate labeled data that can be used to train a downstream NLP model. For instance, an LLM might be prompted to ``\textit{generate 500 movie reviews with positive overall sentiment, and another 500 with negative sentiment}.'' The generated data could then be used to train a binary sentiment classifier, effectively leveraging an LLM as a teacher to a smaller student model. With this demo, we introduce \methodname{}, an open-source Python toolkit for dataset generation. \methodname{} implements common dataset generation workflows, supports a wide range of downstream NLP tasks (such as text classification, question answering, and entity recognition), and is integrated with well-known libraries to facilitate quick experimentation. With \methodname{}, we aim to support researchers in conducting reproducible dataset generation experiments using LLMs and help practitioners apply this approach to train models for downstream tasks.

\end{abstract}

\section{Introduction} \label{sec:introduction}

In recent years, natural language processing (NLP) has witnessed remarkable progress due to the introduction of pre-trained language models (PLMs) \citep{devlin-etal-2019-bert,liu2019roberta,xlm2019lample,he2021deberta}. These PLMs are typically fine-tuned on large human-annotated datasets, resulting in state-of-the-art performance in tasks such as text classification, token classification, and question answering. However, real-world applications of this approach face the bottleneck that sufficient amounts of human-annotated data are often unavailable and too costly to produce manually, especially when domain expertise is required.

\begin{figure}
    \centering
    \includegraphics[width=\linewidth]{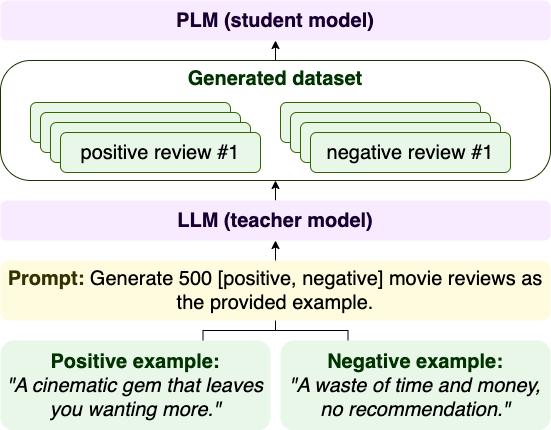}
    \caption{The process of \textit{learning via dataset generation}. A teacher model (LLM) is prompted to generate 500 movie reviews for each sentiment (positive, negative). A smaller student PLM is trained on the generated dataset.}
    \label{fig:overall_process}
\end{figure}

\noindent
\textbf{Dataset generation with teacher LLMs.}
Recently, a paradigm called \textit{zero-shot learning via dataset generation} \citep{meng2022generating,ye-etal-2022-zerogen,ye-etal-2022-progen} has emerged, potentially obviating the need for human-annotated data. This approach leverages the generation capability of large language models (LLMs) to create class-conditioned texts guided by label-descriptive prompts and, optionally, \textit{few-shot} examples of instances of the desired classes. The generated dataset is then used to train a smaller student PLM.


Refer to \Cref{fig:overall_process} for an illustration of this process: In this example, an LLM is instructed to write 500 positive and 500 negative movie reviews. To guide the process, we include an example of a positive and negative review in the prompt. With this prompt and 1-shot example, we generate a dataset of 1,000 movie reviews labeled with binary sentiment. This dataset is used to train a student model to perform binary sentiment analysis.




\noindent
\textbf{Limitations.} However, despite the conceptual simplicity of using LLMs to generate training data, many open questions remain regarding the specifics and ultimate potential of this approach. Questions include: (\textit{1}) How to best prompt the LLM and whether to include examples in the prompt, (\textit{2}) For which downstream NLP task families and specific tasks this approach is effective, and (\textit{3}) Whether it is better to generate large amounts of training data or focus on smaller, high-quality generation efforts. While various current works are investigating these questions for specific tasks, we find that, at present, no open-source library specifically supports research on dataset generation with LLMs.


\begin{figure*}[ht]
    \centering
    \includegraphics[width=\textwidth]{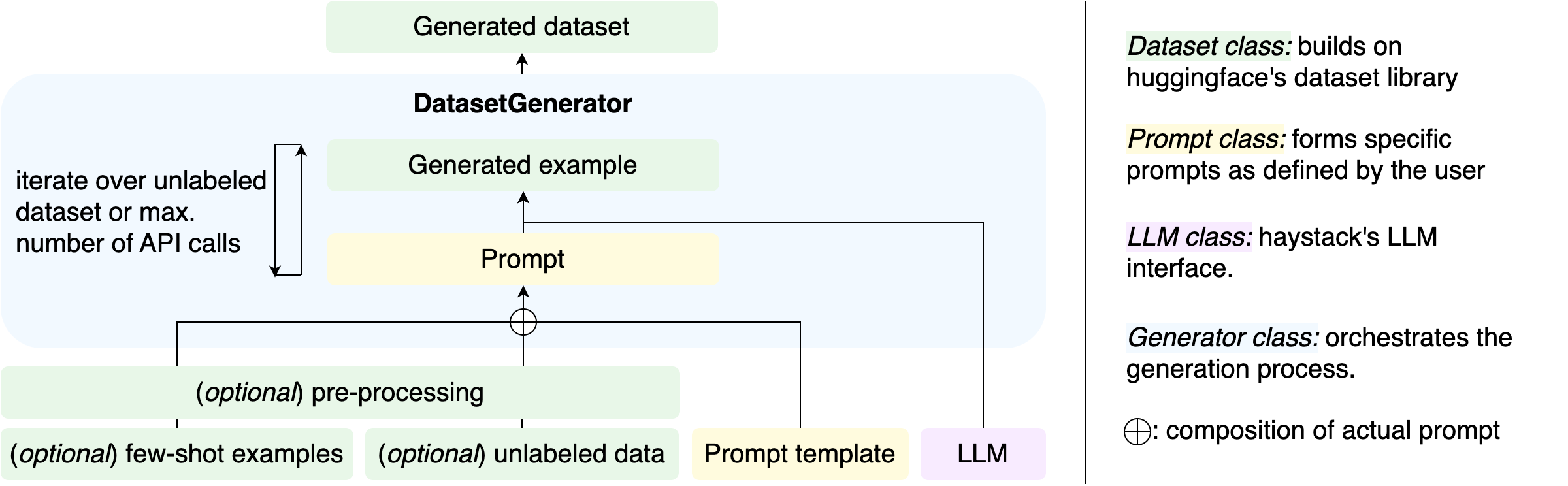}
    \caption{With \methodname{}, the generation process involves a prompt template that creates the final prompt using all provided arguments. The generator class creates training examples until the maximum number of prompt calls is reached, or the unlabeled dataset is fully annotated. Ultimately, the generator class produces a HuggingFace \texttt{Dataset} instance.}
    \label{fig:overview}
\end{figure*}


\noindent
\textbf{Contributions.} To close this gap, we present \methodname{}, an open-source Python library for dataset generation with LLMs.  Our main goals are to facilitate experimentation, enable the application of dataset generation to specific downstream tasks, and encourage the reproducibility of experiments. 

\methodname{} modularizes the dataset generation process and provides a simple interface to facilitate experimentation:  Users may choose which LLM to use, define prompts and label definitions, and leverage existing NLP datasets for few-shot examples and NLP task definitions. Our library includes an integration into HuggingFace's \textsc{datasets} library \citep{lhoest-etal-2021-datasets}, allowing users to easily share generated datasets and use them for training NLP models. We provide examples for various NLP task families, including text classification, textual entailment, question answering, and entity recognition. In this paper:
\begin{itemize}
\item We introduce the \methodname{} library, give an overview of core concepts and usage workflows (\Cref{sec:method}).
\item We present a set of example experiments in which \methodname{} is used to create datasets for various text classification, question answering, and textual entailment tasks (\Cref{sec:experiments}). 
\end{itemize}

We publish the code on GitHub\footnote{\url{https://github.com/flairNLP/fabricator}} under the Apache~2 license.

\section{\methodname{}} \label{sec:method}

We first give a high-level overview of supported generation workflows in 
\methodname{} (\Cref{sec:method-generation_workflows}), discuss the main classes and concepts (\Cref{sec:method-classes_and_concepts}), and walk through an example use case and script (\Cref{sec:method-example_script}).

\subsection{Generation Workflows} \label{sec:method-generation_workflows}


Depending on the downstream task, researchers may have one of three data generation targets we support in \methodname{}:

\begin{description}[leftmargin=0cm]
\item[1. Generate unlabeled data.] The first generation target is to \textit{produce unlabeled data}.  For instance, during the development of a question answering system, we might require a corpus of example questions or a corpus of texts on a particular topic. For this scenario, users provide a prompt $\mathbf{w}$ (such as ``\textit{Generate a text in the domain of history that contains facts someone can ask questions about.}''), and the auto-regressive LLM $G_{\theta}$ generates appropriate text $\mathbf{x}^g$.  



\item[{2. Generate label-conditioned data.}] The second generation target is generating data belonging to a pre-defined class, such as classification tasks. The LLM generates a text $\mathbf{x}^g$ corresponding to a specific label $y$ from a set of labels. 

As discussed in the introduction, one example is to generate training data for a binary sentiment classifier. To achieve this, one must define a set of labels ($\mathbf{y} = \{positive, negative\}$) and a prompt $\mathbf{w}_y$ such as ``\textit{Generate a <$y$> movie review:}.'' The generated sequence $\mathbf{x}^g$ will be paired with the label $y$ to form a training pair $(\mathbf{x}^g, y)$ for fine-tuning.

\item[3. Annotate unlabeled data.] The third generation target holds if an unlabeled text dataset for a domain is already available and only training labels are missing. For instance, a corpus of movie reviews might already be available, but sentiment labels are missing.

In \methodname{}, researchers can add labels to an existing corpus by extending prompt $\mathbf{w}$ with fixed label options $\mathbf{y}$ to form $\mathbf{w}_\mathbf{y}$ like ``\textit{Annotate the movie review either as: positive, negative.}'' The generated label $y$ is then paired with the unlabeled data point $\mathbf{x}^u$ to form a data pair $(\mathbf{x}^u, y)$.
\end{description}

The generation targets defined above will be executed multiple times to generate a corpus of a specified size. The prompt may also be extended to include few-shot examples of each class, as shown in \Cref{fig:overall_process}. The prompt can also handle multiple inputs (for example, for tasks like textual similarity) using pre-defined interfaces in \methodname{}. In all cases, the correct prompt is composed and executed in our backend.

\subsection{Classes and Concepts} \label{sec:method-classes_and_concepts}

As Figure~\ref{fig:overview} illustrates, the key module in our approach is the \texttt{DatasetGenerator} class, which acts as an orchestrator between the LLM (\texttt{PromptNode}), the prompt (\texttt{BasePrompt}), and optionally, the few-shot examples and unlabeled datasets. 

The \texttt{generate()} function within the \texttt{DatasetGenerator} class converts the \texttt{BasePrompt} and the provided few-shot and unlabeled data into a processable prompt for the LLM. The method offers various arguments to steer the generation process. Users can specify parameters like the maximum number of API calls, the sampling strategy of few-shot examples (uniform vs.~stratified), or the number of few-shot examples to use in a single prompt. Our repository contains documentation with details on all available customization options.

\subsubsection{HuggingFace Interoperability through Dataset Class} \label{sec:method-classes-datasets}
\methodname{} operates on the \texttt{Dataset} class from HuggingFace's \textsc{datasets} library. By default, \texttt{generate()} produces the generated data as a \texttt{Dataset} instance. This allows generated datasets to be directly used in existing training scripts of the \textsc{transformers} library \citep{wolf-etal-2020-transformers} and to be shared among researchers via the Huggingface dataset hub.

An existing dataset may also be used as input to the \texttt{generate()} method. Since the \textsc{datasets} library supports a wide range of standard benchmarks and their formats, existing datasets can be easily loaded and used as input. For instance, in some generation workflows, we would like to add labels to an existing corpus or use instances as few-shot examples within a prompt.


\begin{figure*}[ht] 
\begin{lstlisting}[language=Python, caption=A script that uses \methodname{} and generates additional movie reviews based on few-shot examples.]
import os
from datasets import load_dataset
from haystack.nodes import PromptNode
from fabricator import DatasetGenerator, BasePrompt

dataset = load_dataset("processed_fewshot_imdb", split="train")

prompt = BasePrompt(
    task_description="Generate a {} movie review.",
    label_options=["positive", "negative"],
    generate_data_for_column="text",
)

prompt_node = PromptNode(
    model_name_or_path="gpt-3.5-turbo",
    api_key=os.environ.get("OPENAI_API_KEY"),
    max_length=100,
)

generator = DatasetGenerator(prompt_node)
generated_dataset = generator.generate(
    prompt_template=prompt,
    fewshot_dataset=dataset,
    fewshot_sampling_strategy="uniform",
    fewshot_examples_per_class=1,
    fewshot_sampling_column="label",
)
generated_dataset.push_to_hub("generated-movie-reviews")
\end{lstlisting}
\end{figure*}

\subsubsection{Prompt Class} \label{sec:method-classes-prompt}

Prompting is crucial when operating on large language models as it guides the auto-regressive generation process. While in the simplest case, a prompt is a single textual string, we find that many scenarios require more complex prompts and customization options. For instance, when including few-shot examples in a prompt, questions include how many examples to include in each prompt and how these are sampled (uniform vs.~stratified) from available few-shot data across different prompt calls. Similarly, the complexity increases for tasks such as textual entailment (requiring multiple inputs) and entity recognition (potentially requiring transformation of token-level BIOES tags into span-level prompting queries). 

To address these challenges, \methodname{} introduces a simple yet powerful \texttt{BasePrompt} class that offers clear interfaces for customizing prompts for various dataset generation tasks. The interface includes attributes to specify pre-defined label options for label-conditioned generation, and support for having few-shot examples or unlabeled datasets by selecting the relevant columns for generation and few-shot information in the prompt.


Since the prompt class directly operates on the dataset columns, \methodname{} enables a sophisticated and flexible prompt design. To illustrate, when performing a textual similarity task, the user can specify the first sentence and the label as the few-shot information and prompt the LLM to generate a second sentence corresponding to the given sentence and label.

\subsubsection{LLMs} \label{sec:method-classes-llm}
The LLM interface must be stable and ideally compatible with models hosted as APIs or self-hosted LLMs. We leverage the \textsc{Haystack}\footnote{\url{https://github.com/deepset-ai/Haystack}} framework \citep{Pietsch_Haystack_the_end-to-end_2019}, specifically the \texttt{PromptNode} class, for interactions with LLMs. The \texttt{PromptNode} implementation allows users to select and use LLMs from various model providers, including HuggingFace, OpenAI, Azure, Anthropic, and Cohere.

\renewcommand{\arraystretch}{1.1}
\begin{table*}[ht]
\centering
\begin{tabular}{lccccc}
\toprule
\textbf{Dataset} & \textbf{Labels} & \multicolumn{4}{c}{\textbf{\# Training examples}} \\
& & 50 & 500 & 1k & all (max. 10k) \\
\midrule
\multirow{2}{*}{IMDB} & Gold & $37.6 \pm 35.8$ & $88.5 \pm 0.8$ & $90.0 \pm 0.4$ & $93.0 \pm 0.2$\\
& Generated & $\textbf{53.8} \pm 11.5$ & $\textbf{88.8} \pm 0.6$ & $\textbf{90.2} \pm 0.4$ & $\textbf{92.0} \pm 0.1$\\ \cline{2-6}
\multirow{2}{*}{MRPC} & Gold & $66.6 \pm 0.8$ & $73.0 \pm 1.3$ & $75.2 \pm 1.1$ & $83.9 \pm 0.2$\\
& Generated & $\textbf{68.4} \pm 0.8$ & $\textbf{72.1} \pm  1.0$ & $72.4 \pm 1.2$ & $75.8 \pm 0.7$ \\\cline{2-6}
\multirow{2}{*}{SNLI} & Gold & $38.5 \pm 2.5$ & $64.7 \pm 0.9$ & $71.3 \pm 0.7$ & $82.1 \pm 0.4$ \\
& Generated & $\textbf{42.2} \pm 2.4$ & $54.8 \pm 1.0$ & $56.1 \pm 1.1$ & $63.1 \pm 0.7$ \\\cline{2-6}
\multirow{2}{*}{TREC-6} & Gold & $50.4 \pm 7.6$ & $93.6 \pm 0.6$ & $94.9 \pm 1.1$ & $97.5 \pm 0.4$ \\
& Generated & $39.8 \pm 4.5$ & $79.3 \pm 2.2$ & $80.8 \pm 3.0$ & $82.4 \pm 1.1$ \\\cline{2-6}
\multirow{2}{*}{SQuAD} & Gold & -& -& $39.1 \pm 4.9$& $68.8 \pm 0.5$\\
& Generated & -& -& $\textbf{46.8} \pm 1.1$& $52.5 \pm 0.3$\\
\bottomrule
\end{tabular}
\caption{Results on re-annotation experiments using 2 few-shot examples per prompt (uniformly sampled from 6 few-shot examples per class). We report accuracy except for SQuAD, where we report F1, and highlight bold those experiments where generated data yielded similar scores as human-annotated data. We observe that GPT-3.5 is not able to annotate on human-level performance except for simple classification tasks such as IMDB.}
\label{tab:experiment1-dataset_annotation_results}
\end{table*}

\subsection{Example Script} \label{sec:method-example_script}


In Listing 1, we introduce an example script in which \methodname{} is used to generate additional movie reviews for training a binary sentiment classification model (refer to generation workflow 2 as defined in \Cref{sec:method-generation_workflows}). 
To implement this, we define: 
\begin{itemize}
    \item a pre-processed few-shot dataset (\texttt{dataset}, line 6) having labels in natural language form (e.g., 0 becomes ``negative''). These examples are used to augment the generation prompt,
    \item a prompt template (\texttt{prompt}, line 8) specifying the instruction to the LLM,
    \item an LLM to use as teacher model (\texttt{prompt\_node}, line 14), 
    \item a \texttt{DatasetGenerator} to execute the generation process with all parameters (\texttt{generator}, line 20).  
\end{itemize}



The prompt is configured in the constructor of the \texttt{BasePrompt} class (lines 8-12): We set a \texttt{task\_description} with a placeholder for \texttt{label\_options} that we provide as a separate argument. We also specify for which column in the loaded dataset to predict labels.


We then define a teacher LLM (lines 14-18) and pass datasets, prompt, and LLM to the \texttt{DatasetGenerator} orchestrator class (lines 20-27). Here, we specify a few-shot strategy to sample one label from the ``label'' column uniformly during generation. We do so to generate either a positive or a negative review. Upon completion, the \texttt{generate{}} function returns the annotated \texttt{Dataset} instance.

\section{Experiments}
\label{sec:experiments}

To illustrate how  \methodname{} could be used in research, we conduct an exploratory evaluation of two scenarios: (\textit{1}) how models trained on generated datasets compare to models trained on human-annotated datasets, and (\textit{2}) whether few-shot examples in the prompt improve generated datasets. 


To do so, we train smaller PLMs on generated datasets and evaluate them on the human-labeled test split of the respective benchmark. For question answering, we fine-tune a \texttt{roberta-base} PLM~\cite{liu2019roberta}. For all other tasks, we fine-tune a \texttt{bert-base-uncased} PLM~\cite{devlin-etal-2019-bert}. The hyperparameters are listed in \Cref{sec:appendix-hyperparameters}. We report the score and standard deviation averaged over 5 random seeds for each experiment.



\begin{table*}[ht]
\centering
\begin{tabular}{lcccccc}
\toprule
\textbf{Dataset} & \textbf{\# few-shot examples} & \multicolumn{5}{c}{\textbf{\# examples per class used in prompt}} \\
& \textbf{per class} & 0 & 1 & 2 & 3 & 4 \\
\midrule
\multirow{5}{*}{TREC-6} & 0 & $45.5 \pm 2.3$ & - & - & - & - \\
& 2 & - & $70.0 \pm 1.6$ & $65.5 \pm 0.9$ & - & - \\
& 4 & - & $79.5 \pm 1.1$ & $71.1 \pm 2.0$ & $\textbf{86.6} \pm 0.6$ & $69.8 \pm 1.5$ \\
& 8 & - & $76.1 \pm 1.9$ & $\textbf{79.5} \pm 1.3$ & $\textbf{81.0} \pm 1.8$ & $\textbf{87.4} \pm 0.6$ \\
& 16 & - & $72.7 \pm 2.1$ & $78.1 \pm 1.9$ & $\textbf{81.0} \pm 2.4$ & $74.2 \pm 1.4$ \\
\bottomrule
\end{tabular}
\caption{Results on 500 annotated TREC-6 examples using varying amounts of few-shot examples. We sweep over the number of few-shot examples and the number of few-shot examples used in the actual prompt. We highlight bold where increasing few-shot examples improves over the $79.3$ TREC-6 score of Experiment 1 (Table~\ref{tab:experiment1-dataset_annotation_results}). 
}
\label{tab:experiment2-hyperparameter_annotation_results}
\end{table*}

\subsection{Experiment 1: Comparison of Generated and Human-Annotated Datasets} \label{sec:experiment1}

We re-annotate existing benchmark datasets with generated labels in the first experiment. This experiment aims to measure the difference in accuracy of downstream task models trained on human-annotated data compared to models trained on generated data. We evaluate text classification, textual similarity, and extractive question answering tasks.

\noindent 
\textbf{Experimental setup.} We conduct this evaluation on 5 datasets spanning 3 NLP tasks: We use IMDB \citep{maas-etal-2011-learning}, a binary sentiment classification benchmark, and TREC-6 \citep{li-roth-2002-learning}, a 6-class question type categorization dataset to evaluate text classification tasks. We  use the 2-class MRPC \citep{dolan-brockett-2005-automatically} and the 3-class SNLI \citep{bowman-etal-2015-large}) datasets to evaluate textual similarity tasks. Finally, we use SQuAD-v2 \citep{rajpurkar-etal-2016-squad}) to evaluate extractive question answering. We use generation prompts augmented by 2 examples per prompt sampled from 6 possible few-shot examples per class.  

\noindent 
\textbf{Results (Table~\ref{tab:experiment1-dataset_annotation_results}).} For all datasets, we compare a generated dataset of 50, 500, 1k and the full dataset (limited to 10k if it is larger) to gold-annotated data of the same size. For question answering, models need to be trained on at least 1k to obtain representative results, so we do not report scores for 50 or 500 examples for SQuAD.

We find that for simple tasks such as binary sentiment classification (IMDB), models trained on the annotations by LLMs achieve similar accuracy on the gold-labeled test split ($\downarrow$1.0 pp. in accuracy with 10k training examples). However, we as the complexity of datasets increases (text classification with more classes and extractive question answering), we observe that the performance of models trained on LLM-annotated datasets falls short ($\downarrow$19.0 pp.~for SNLI and $\downarrow$16.3 pp.~for SQuAD, with 10k training examples). 

These performance gaps indicate that the usefulness of LLMs as teacher models depends on the specific task. In the next section, we present an experiment that explores how to close this gap by using additional few-shot examples.


\subsection{Experiment 2: Impact of Few-Shot Examples} \label{sec:experiment2}
In the second example experiment, we re-annotate TREC-6 using a varying number of few-shot examples. This experiment aims to determine whether adding few-shot examples for each class improves dataset generation with \methodname{}. We investigate two variables: (\textit{1}) The total number of available few-shot examples per class and (\textit{2}) the actual number of few-shot examples included per prompt. For instance, there might be 8 few-shot examples available in total, but only 3 are randomly sampled to be included in each prompt call.

\noindent 
\textbf{Results (Table~\ref{tab:experiment2-hyperparameter_annotation_results}).} 
We note a generally positive trend in that increasing the number of available few-shot examples (column \textit{\# few-shot examples per class}) and increasing the number of examples used in each prompt (column \textit{\# examples per class used in prompt}) improves model performance. In particular, we find many settings that outperform the numbers of our previous experiment (where we sampled 2 examples per prompt out of a total of 6 possible examples), highlighted bold in Table~\ref{tab:experiment2-hyperparameter_annotation_results}.

However, we also find that improvements become uneven when \textit{\# examples per class used in prompt} is increased above 3, indicating prompts should not be overloaded with too many examples.

\section{Related Work} \label{sec:related_work}

Significant progress has been achieved in enhancing dataset generation with teacher LLMs \citep{schick-schutze-2021-generating,meng2022generating,ye-etal-2022-zerogen,inpars2022bonifacio,peng2023instruction,pmlr-v202-meng23b}, effectively selecting few-shot examples \citep{liu-etal-2022-makes,gunasekar2023textbooks} and assessing the quality of datasets produced by LLMs \citep{Gilardi_2023,chen-etal-2023-empirical}. However, we note a lack of accessible frameworks that facilitate straightforward and reproducible dataset generation using teacher LLMs. While existing open-source toolkits like OpenPrompt \citep{ding-etal-2022-openprompt} partially extend to dataset generation scenarios, our approach stands apart by having lightweight, dedicated interfaces for the introduced generation tasks, supporting a wide range of LLMs using haystack, and integrating with HuggingFace \textsc{datasets} for easy evaluation.

Prompt-based learning \citep{liu2021gpt,gao-etal-2021-making,schick-schutze-2021-exploiting,le-scao-rush-2021-many} is another line of research that has proven useful in improving downstream tasks in zero- and few-shot settings by leveraging LLMs' pre-training objectives \cite{brown2020language,ouyang2022rlhf,zhang2022opt,workshop2023bloom,touvron2023llama}. However, the availability of training data in low-resource scenarios is still crucial \citep{perez2021true,sahu-etal-2022-data}. Therefore, our method also seeks to fill this gap by providing a comprehensive and easily reproducible dataset generation toolkit.

\section{Conclusion} \label{sec:conclusion}
We introduced \methodname{}, a user-friendly library for dataset generation utilizing LLMs. With \methodname{}, researchers access a highly customizable interface that enables efficient research on zero-shot and few-shot learning via dataset generation. Further, we implemented various baselines using generated datasets to illustrate potential applications of our repository and plan to support further downstream tasks in the future. We believe that \methodname{} will be a valuable tool for the NLP community, facilitating advancements in dataset generation and fostering research in various natural language processing domains.

\clearpage

\section*{Limitations}

While our paper aims to address dataset creation for a wide range of downstream tasks, it is important to acknowledge certain limitations in our study. Firstly, during our repository's evaluation phase, we could only test and assess a subset of tasks due to resource and time constraints. Our evaluation may only cover a portion of the tasks researchers and practitioners commonly encounter in their work. Future work must expand the evaluation to include a broader range of tasks to provide a more comprehensive understanding of the repository's effectiveness.

Additionally, despite our best efforts in designing the repository layout to be versatile and adaptable, there might be specific tasks or domains where our repository's structure or features may not be directly applicable. We acknowledge that the landscape of downstream tasks is diverse and constantly evolving, which may require tailored approaches or extensions to our existing framework. Further, we aim to include existing research targeting high-quality dataset generation (e.g., \citet{ye-etal-2022-progen}) and conduct our own research on quality and diversity metrics to steer the generation process. We encourage open-source contributions and active engagement from the community to address these limitations. By involving a more comprehensive range of perspectives and expertise, we aim to consistently improve the repository and enhance its suitability for various task requirements.

Furthermore, while we have endeavored to provide thorough documentation and guidelines within the repository, there is always a possibility of overlooked issues or unforeseen challenges that may arise during dataset creation.

\section*{Ethics Statement}

While large language models have shown remarkable advancements in natural language understanding and generation, their capabilities also raise important ethical considerations. One prominent concern is the potential for hallucination, where the models may generate false or misleading information. This aspect can have serious implications, especially when datasets are created for critical domains such as medicine, law, or journalism. It is crucial to exercise caution and verify the accuracy and reliability of outputs generated by our repository, particularly when making decisions that have real-world consequences.

Another ethical concern is the presence of biases in language models, which can perpetuate and amplify societal prejudices and inequalities. These biases can arise from biased training data~\cite{haller2023opiniongpt} or biased patterns in human-generated text that the models learn from. Since our repository is in an early stage, we emphasize to carefully inspect created datasets to identify and rectify biases that may be present.

To ensure a responsible dataset creation process, it is essential to engage in thorough data validation, including identifying and addressing potential biases, checking data sources for reliability and credibility, and involving diverse perspectives in dataset collection and annotation processes. Moreover, continuous monitoring and auditing of the models' outputs and performance can help identify and rectify any ethical concerns arising during deployment.

\section*{Acknowledgements}
We thank all reviewers for their valuable comments. Jonas Golde is supported by the German Federal Ministry of Economic Affairs and Climate Action (BMWK) as part of the project ENA (KK5148001LB0). Alan Akbik and Patrick Haller are supported by the Deutsche Forschungsgemeinschaft (DFG, German Research Foundation) under Emmy Noether grant ``Eidetic Representations of Natural Language'' (project number 448414230). Alan Akbik is furthermore supported under 
Germany’s Excellence Strategy "Science of Intelligence" (EXC 2002/1, project number 390523135). Felix Hamborg is supported by the WIN program of the Heidelberg Academy of Sciences and Humanities, financed by the Ministry of Science, Research and Arts of the State of Baden-Wurttemberg, Germany.

\bibliography{anthology,custom}

\begin{thebibliography}{38}
\expandafter\ifx\csname natexlab\endcsname\relax\def\natexlab#1{#1}\fi

\bibitem[{Bonifacio et~al.(2022)Bonifacio, Abonizio, Fadaee, and Nogueira}]{inpars2022bonifacio}
Luiz Bonifacio, Hugo Abonizio, Marzieh Fadaee, and Rodrigo Nogueira. 2022.
\newblock \href {https://doi.org/10.1145/3477495.3531863} {Inpars: Unsupervised dataset generation for information retrieval}.
\newblock In \emph{Proceedings of the 45th International ACM SIGIR Conference on Research and Development in Information Retrieval}, SIGIR '22, page 2387–2392, New York, NY, USA. Association for Computing Machinery.

\bibitem[{Bowman et~al.(2015)Bowman, Angeli, Potts, and Manning}]{bowman-etal-2015-large}
Samuel~R. Bowman, Gabor Angeli, Christopher Potts, and Christopher~D. Manning. 2015.
\newblock \href {https://doi.org/10.18653/v1/D15-1075} {A large annotated corpus for learning natural language inference}.
\newblock In \emph{Proceedings of the 2015 Conference on Empirical Methods in Natural Language Processing}, pages 632--642, Lisbon, Portugal. Association for Computational Linguistics.

\bibitem[{Brown et~al.(2020)Brown, Mann, Ryder, Subbiah, Kaplan, Dhariwal, Neelakantan, Shyam, Sastry, Askell, Agarwal, Herbert-Voss, Krueger, Henighan, Child, Ramesh, Ziegler, Wu, Winter, Hesse, Chen, Sigler, Litwin, Gray, Chess, Clark, Berner, McCandlish, Radford, Sutskever, and Amodei}]{brown2020language}
Tom Brown, Benjamin Mann, Nick Ryder, Melanie Subbiah, Jared~D Kaplan, Prafulla Dhariwal, Arvind Neelakantan, Pranav Shyam, Girish Sastry, Amanda Askell, Sandhini Agarwal, Ariel Herbert-Voss, Gretchen Krueger, Tom Henighan, Rewon Child, Aditya Ramesh, Daniel Ziegler, Jeffrey Wu, Clemens Winter, Chris Hesse, Mark Chen, Eric Sigler, Mateusz Litwin, Scott Gray, Benjamin Chess, Jack Clark, Christopher Berner, Sam McCandlish, Alec Radford, Ilya Sutskever, and Dario Amodei. 2020.
\newblock \href {https://proceedings.neurips.cc/paper_files/paper/2020/file/1457c0d6bfcb4967418bfb8ac142f64a-Paper.pdf} {Language models are few-shot learners}.
\newblock In \emph{Advances in Neural Information Processing Systems}, volume~33, pages 1877--1901. Curran Associates, Inc.

\bibitem[{Chen et~al.(2023)Chen, Tam, Raffel, Bansal, and Yang}]{chen-etal-2023-empirical}
Jiaao Chen, Derek Tam, Colin Raffel, Mohit Bansal, and Diyi Yang. 2023.
\newblock \href {https://doi.org/10.1162/tacl_a_00542} {An empirical survey of data augmentation for limited data learning in {NLP}}.
\newblock \emph{Transactions of the Association for Computational Linguistics}, 11:191--211.

\bibitem[{Conneau and Lample(2019)}]{xlm2019lample}
Alexis Conneau and Guillaume Lample. 2019.
\newblock \href {https://proceedings.neurips.cc/paper_files/paper/2019/file/c04c19c2c2474dbf5f7ac4372c5b9af1-Paper.pdf} {Cross-lingual language model pretraining}.
\newblock In \emph{Advances in Neural Information Processing Systems}, volume~32. Curran Associates, Inc.

\bibitem[{Devlin et~al.(2019)Devlin, Chang, Lee, and Toutanova}]{devlin-etal-2019-bert}
Jacob Devlin, Ming-Wei Chang, Kenton Lee, and Kristina Toutanova. 2019.
\newblock \href {https://doi.org/10.18653/v1/N19-1423} {{BERT}: Pre-training of deep bidirectional transformers for language understanding}.
\newblock In \emph{Proceedings of the 2019 Conference of the North {A}merican Chapter of the Association for Computational Linguistics: Human Language Technologies, Volume 1 (Long and Short Papers)}, pages 4171--4186, Minneapolis, Minnesota. Association for Computational Linguistics.

\bibitem[{Ding et~al.(2022)Ding, Hu, Zhao, Chen, Liu, Zheng, and Sun}]{ding-etal-2022-openprompt}
Ning Ding, Shengding Hu, Weilin Zhao, Yulin Chen, Zhiyuan Liu, Haitao Zheng, and Maosong Sun. 2022.
\newblock \href {https://doi.org/10.18653/v1/2022.acl-demo.10} {{O}pen{P}rompt: An open-source framework for prompt-learning}.
\newblock In \emph{Proceedings of the 60th Annual Meeting of the Association for Computational Linguistics: System Demonstrations}, pages 105--113, Dublin, Ireland. Association for Computational Linguistics.

\bibitem[{Dolan and Brockett(2005)}]{dolan-brockett-2005-automatically}
William~B. Dolan and Chris Brockett. 2005.
\newblock \href {https://aclanthology.org/I05-5002} {Automatically constructing a corpus of sentential paraphrases}.
\newblock In \emph{Proceedings of the Third International Workshop on Paraphrasing ({IWP}2005)}.

\bibitem[{Gao et~al.(2021)Gao, Fisch, and Chen}]{gao-etal-2021-making}
Tianyu Gao, Adam Fisch, and Danqi Chen. 2021.
\newblock \href {https://doi.org/10.18653/v1/2021.acl-long.295} {Making pre-trained language models better few-shot learners}.
\newblock In \emph{Proceedings of the 59th Annual Meeting of the Association for Computational Linguistics and the 11th International Joint Conference on Natural Language Processing (Volume 1: Long Papers)}, pages 3816--3830, Online. Association for Computational Linguistics.

\bibitem[{Gilardi et~al.(2023)Gilardi, Alizadeh, and Kubli}]{Gilardi_2023}
Fabrizio Gilardi, Meysam Alizadeh, and Maël Kubli. 2023.
\newblock \href {https://doi.org/10.1073/pnas.2305016120} {{ChatGPT} outperforms crowd workers for text-annotation tasks}.
\newblock \emph{Proceedings of the National Academy of Sciences}, 120(30).

\bibitem[{Gunasekar et~al.(2023)Gunasekar, Zhang, Aneja, Mendes, Giorno, Gopi, Javaheripi, Kauffmann, de~Rosa, Saarikivi, Salim, Shah, Behl, Wang, Bubeck, Eldan, Kalai, Lee, and Li}]{gunasekar2023textbooks}
Suriya Gunasekar, Yi~Zhang, Jyoti Aneja, Caio César~Teodoro Mendes, Allie~Del Giorno, Sivakanth Gopi, Mojan Javaheripi, Piero Kauffmann, Gustavo de~Rosa, Olli Saarikivi, Adil Salim, Shital Shah, Harkirat~Singh Behl, Xin Wang, Sébastien Bubeck, Ronen Eldan, Adam~Tauman Kalai, Yin~Tat Lee, and Yuanzhi Li. 2023.
\newblock \href {http://arxiv.org/abs/2306.11644} {Textbooks are all you need}.

\bibitem[{Haller et~al.(2023)Haller, Aynetdinov, and Akbik}]{haller2023opiniongpt}
Patrick Haller, Ansar Aynetdinov, and Alan Akbik. 2023.
\newblock \href {http://arxiv.org/abs/2309.03876} {Opiniongpt: Modelling explicit biases in instruction-tuned llms}.

\bibitem[{He et~al.(2021)He, Liu, Gao, and Chen}]{he2021deberta}
Pengcheng He, Xiaodong Liu, Jianfeng Gao, and Weizhu Chen. 2021.
\newblock \href {https://openreview.net/forum?id=XPZIaotutsD} {Deberta: Decoding-enhanced bert with disentangled attention}.
\newblock In \emph{International Conference on Learning Representations}.

\bibitem[{Le~Scao and Rush(2021)}]{le-scao-rush-2021-many}
Teven Le~Scao and Alexander Rush. 2021.
\newblock \href {https://doi.org/10.18653/v1/2021.naacl-main.208} {How many data points is a prompt worth?}
\newblock In \emph{Proceedings of the 2021 Conference of the North American Chapter of the Association for Computational Linguistics: Human Language Technologies}, pages 2627--2636, Online. Association for Computational Linguistics.

\bibitem[{Lhoest et~al.(2021)Lhoest, Villanova~del Moral, Jernite, Thakur, von Platen, Patil, Chaumond, Drame, Plu, Tunstall, Davison, {\v{S}}a{\v{s}}ko, Chhablani, Malik, Brandeis, Le~Scao, Sanh, Xu, Patry, McMillan-Major, Schmid, Gugger, Delangue, Matussi{\`e}re, Debut, Bekman, Cistac, Goehringer, Mustar, Lagunas, Rush, and Wolf}]{lhoest-etal-2021-datasets}
Quentin Lhoest, Albert Villanova~del Moral, Yacine Jernite, Abhishek Thakur, Patrick von Platen, Suraj Patil, Julien Chaumond, Mariama Drame, Julien Plu, Lewis Tunstall, Joe Davison, Mario {\v{S}}a{\v{s}}ko, Gunjan Chhablani, Bhavitvya Malik, Simon Brandeis, Teven Le~Scao, Victor Sanh, Canwen Xu, Nicolas Patry, Angelina McMillan-Major, Philipp Schmid, Sylvain Gugger, Cl{\'e}ment Delangue, Th{\'e}o Matussi{\`e}re, Lysandre Debut, Stas Bekman, Pierric Cistac, Thibault Goehringer, Victor Mustar, Fran{\c{c}}ois Lagunas, Alexander Rush, and Thomas Wolf. 2021.
\newblock \href {https://doi.org/10.18653/v1/2021.emnlp-demo.21} {Datasets: A community library for natural language processing}.
\newblock In \emph{Proceedings of the 2021 Conference on Empirical Methods in Natural Language Processing: System Demonstrations}, pages 175--184, Online and Punta Cana, Dominican Republic. Association for Computational Linguistics.

\bibitem[{Li and Roth(2002)}]{li-roth-2002-learning}
Xin Li and Dan Roth. 2002.
\newblock \href {https://aclanthology.org/C02-1150} {Learning question classifiers}.
\newblock In \emph{{COLING} 2002: The 19th International Conference on Computational Linguistics}.

\bibitem[{Liu et~al.(2022)Liu, Shen, Zhang, Dolan, Carin, and Chen}]{liu-etal-2022-makes}
Jiachang Liu, Dinghan Shen, Yizhe Zhang, Bill Dolan, Lawrence Carin, and Weizhu Chen. 2022.
\newblock \href {https://doi.org/10.18653/v1/2022.deelio-1.10} {What makes good in-context examples for {GPT}-3?}
\newblock In \emph{Proceedings of Deep Learning Inside Out (DeeLIO 2022): The 3rd Workshop on Knowledge Extraction and Integration for Deep Learning Architectures}, pages 100--114, Dublin, Ireland and Online. Association for Computational Linguistics.

\bibitem[{Liu et~al.(2021)Liu, Zheng, Du, Ding, Qian, Yang, and Tang}]{liu2021gpt}
Xiao Liu, Yanan Zheng, Zhengxiao Du, Ming Ding, Yujie Qian, Zhilin Yang, and Jie Tang. 2021.
\newblock \href {http://arxiv.org/abs/2103.10385} {Gpt understands, too}.

\bibitem[{Liu et~al.(2019)Liu, Ott, Goyal, Du, Joshi, Chen, Levy, Lewis, Zettlemoyer, and Stoyanov}]{liu2019roberta}
Yinhan Liu, Myle Ott, Naman Goyal, Jingfei Du, Mandar Joshi, Danqi Chen, Omer Levy, Mike Lewis, Luke Zettlemoyer, and Veselin Stoyanov. 2019.
\newblock \href {http://arxiv.org/abs/1907.11692} {Roberta: A robustly optimized bert pretraining approach}.

\bibitem[{Loshchilov and Hutter(2019)}]{loshchilov2018decoupled}
Ilya Loshchilov and Frank Hutter. 2019.
\newblock \href {https://openreview.net/forum?id=Bkg6RiCqY7} {Decoupled weight decay regularization}.
\newblock In \emph{International Conference on Learning Representations}.

\bibitem[{Maas et~al.(2011)Maas, Daly, Pham, Huang, Ng, and Potts}]{maas-etal-2011-learning}
Andrew~L. Maas, Raymond~E. Daly, Peter~T. Pham, Dan Huang, Andrew~Y. Ng, and Christopher Potts. 2011.
\newblock \href {https://aclanthology.org/P11-1015} {Learning word vectors for sentiment analysis}.
\newblock In \emph{Proceedings of the 49th Annual Meeting of the Association for Computational Linguistics: Human Language Technologies}, pages 142--150, Portland, Oregon, USA. Association for Computational Linguistics.

\bibitem[{Meng et~al.(2022)Meng, Huang, Zhang, and Han}]{meng2022generating}
Yu~Meng, Jiaxin Huang, Yu~Zhang, and Jiawei Han. 2022.
\newblock \href {https://proceedings.neurips.cc/paper_files/paper/2022/file/0346c148ba1c21c6b4780a961ea141dc-Paper-Conference.pdf} {Generating training data with language models: Towards zero-shot language understanding}.
\newblock In \emph{Advances in Neural Information Processing Systems}, volume~35, pages 462--477. Curran Associates, Inc.

\bibitem[{Meng et~al.(2023)Meng, Michalski, Huang, Zhang, Abdelzaher, and Han}]{pmlr-v202-meng23b}
Yu~Meng, Martin Michalski, Jiaxin Huang, Yu~Zhang, Tarek Abdelzaher, and Jiawei Han. 2023.
\newblock \href {https://proceedings.mlr.press/v202/meng23b.html} {Tuning language models as training data generators for augmentation-enhanced few-shot learning}.
\newblock In \emph{Proceedings of the 40th International Conference on Machine Learning}, volume 202 of \emph{Proceedings of Machine Learning Research}, pages 24457--24477. PMLR.

\bibitem[{Ouyang et~al.(2022)Ouyang, Wu, Jiang, Almeida, Wainwright, Mishkin, Zhang, Agarwal, Slama, Ray, Schulman, Hilton, Kelton, Miller, Simens, Askell, Welinder, Christiano, Leike, and Lowe}]{ouyang2022rlhf}
Long Ouyang, Jeffrey Wu, Xu~Jiang, Diogo Almeida, Carroll Wainwright, Pamela Mishkin, Chong Zhang, Sandhini Agarwal, Katarina Slama, Alex Ray, John Schulman, Jacob Hilton, Fraser Kelton, Luke Miller, Maddie Simens, Amanda Askell, Peter Welinder, Paul~F Christiano, Jan Leike, and Ryan Lowe. 2022.
\newblock \href {https://proceedings.neurips.cc/paper_files/paper/2022/file/b1efde53be364a73914f58805a001731-Paper-Conference.pdf} {Training language models to follow instructions with human feedback}.
\newblock In \emph{Advances in Neural Information Processing Systems}, volume~35, pages 27730--27744. Curran Associates, Inc.

\bibitem[{Peng et~al.(2023)Peng, Li, He, Galley, and Gao}]{peng2023instruction}
Baolin Peng, Chunyuan Li, Pengcheng He, Michel Galley, and Jianfeng Gao. 2023.
\newblock \href {http://arxiv.org/abs/2304.03277} {Instruction tuning with gpt-4}.

\bibitem[{Perez et~al.(2021)Perez, Kiela, and Cho}]{perez2021true}
Ethan Perez, Douwe Kiela, and Kyunghyun Cho. 2021.
\newblock \href {https://proceedings.neurips.cc/paper_files/paper/2021/file/5c04925674920eb58467fb52ce4ef728-Paper.pdf} {True few-shot learning with language models}.
\newblock In \emph{Advances in Neural Information Processing Systems}, volume~34, pages 11054--11070. Curran Associates, Inc.

\bibitem[{Pietsch et~al.(2019)Pietsch, Möller, Kostic, Risch, Pippi, Jobanputra, Zanzottera, Cerza, Blagojevic, Stadelmann, Soni, and Lee}]{Pietsch_Haystack_the_end-to-end_2019}
Malte Pietsch, Timo Möller, Bogdan Kostic, Julian Risch, Massimiliano Pippi, Mayank Jobanputra, Sara Zanzottera, Silvano Cerza, Vladimir Blagojevic, Thomas Stadelmann, Tanay Soni, and Sebastian Lee. 2019.
\newblock \href {https://github.com/deepset-ai/haystack} {{Haystack: the end-to-end NLP framework for pragmatic builders}}.

\bibitem[{Rajpurkar et~al.(2016)Rajpurkar, Zhang, Lopyrev, and Liang}]{rajpurkar-etal-2016-squad}
Pranav Rajpurkar, Jian Zhang, Konstantin Lopyrev, and Percy Liang. 2016.
\newblock \href {https://doi.org/10.18653/v1/D16-1264} {{SQ}u{AD}: 100,000+ questions for machine comprehension of text}.
\newblock In \emph{Proceedings of the 2016 Conference on Empirical Methods in Natural Language Processing}, pages 2383--2392, Austin, Texas. Association for Computational Linguistics.

\bibitem[{Sahu et~al.(2022)Sahu, Rodriguez, Laradji, Atighehchian, Vazquez, and Bahdanau}]{sahu-etal-2022-data}
Gaurav Sahu, Pau Rodriguez, Issam Laradji, Parmida Atighehchian, David Vazquez, and Dzmitry Bahdanau. 2022.
\newblock \href {https://doi.org/10.18653/v1/2022.nlp4convai-1.5} {Data augmentation for intent classification with off-the-shelf large language models}.
\newblock In \emph{Proceedings of the 4th Workshop on NLP for Conversational AI}, pages 47--57, Dublin, Ireland. Association for Computational Linguistics.

\bibitem[{Scao et~al.(2023)Scao, Fan, Akiki, Pavlick, Ilić, Hesslow, Castagné, Luccioni, Yvon, and et~al.}]{workshop2023bloom}
Teven~Le Scao, Angela Fan, Christopher Akiki, Ellie Pavlick, Suzana Ilić, Daniel Hesslow, Roman Castagné, Alexandra~Sasha Luccioni, François Yvon, and Matthias~Gallé et~al. 2023.
\newblock \href {http://arxiv.org/abs/2211.05100} {Bloom: A 176b-parameter open-access multilingual language model}.

\bibitem[{Schick and Sch{\"u}tze(2021{\natexlab{a}})}]{schick-schutze-2021-exploiting}
Timo Schick and Hinrich Sch{\"u}tze. 2021{\natexlab{a}}.
\newblock \href {https://doi.org/10.18653/v1/2021.eacl-main.20} {Exploiting cloze-questions for few-shot text classification and natural language inference}.
\newblock In \emph{Proceedings of the 16th Conference of the European Chapter of the Association for Computational Linguistics: Main Volume}, pages 255--269, Online. Association for Computational Linguistics.

\bibitem[{Schick and Sch{\"u}tze(2021{\natexlab{b}})}]{schick-schutze-2021-generating}
Timo Schick and Hinrich Sch{\"u}tze. 2021{\natexlab{b}}.
\newblock \href {https://doi.org/10.18653/v1/2021.emnlp-main.555} {Generating datasets with pretrained language models}.
\newblock In \emph{Proceedings of the 2021 Conference on Empirical Methods in Natural Language Processing}, pages 6943--6951, Online and Punta Cana, Dominican Republic. Association for Computational Linguistics.

\bibitem[{Tjong Kim~Sang and De~Meulder(2003)}]{tjong-kim-sang-de-meulder-2003-introduction}
Erik~F. Tjong Kim~Sang and Fien De~Meulder. 2003.
\newblock \href {https://aclanthology.org/W03-0419} {Introduction to the {C}o{NLL}-2003 shared task: Language-independent named entity recognition}.
\newblock In \emph{Proceedings of the Seventh Conference on Natural Language Learning at {HLT}-{NAACL} 2003}, pages 142--147.

\bibitem[{Touvron et~al.(2023)Touvron, Lavril, Izacard, Martinet, Lachaux, Lacroix, Rozière, Goyal, Hambro, Azhar, Rodriguez, Joulin, Grave, and Lample}]{touvron2023llama}
Hugo Touvron, Thibaut Lavril, Gautier Izacard, Xavier Martinet, Marie-Anne Lachaux, Timothée Lacroix, Baptiste Rozière, Naman Goyal, Eric Hambro, Faisal Azhar, Aurelien Rodriguez, Armand Joulin, Edouard Grave, and Guillaume Lample. 2023.
\newblock \href {http://arxiv.org/abs/2302.13971} {Llama: Open and efficient foundation language models}.

\bibitem[{Wolf et~al.(2020)Wolf, Debut, Sanh, Chaumond, Delangue, Moi, Cistac, Rault, Louf, Funtowicz, Davison, Shleifer, von Platen, Ma, Jernite, Plu, Xu, Le~Scao, Gugger, Drame, Lhoest, and Rush}]{wolf-etal-2020-transformers}
Thomas Wolf, Lysandre Debut, Victor Sanh, Julien Chaumond, Clement Delangue, Anthony Moi, Pierric Cistac, Tim Rault, Remi Louf, Morgan Funtowicz, Joe Davison, Sam Shleifer, Patrick von Platen, Clara Ma, Yacine Jernite, Julien Plu, Canwen Xu, Teven Le~Scao, Sylvain Gugger, Mariama Drame, Quentin Lhoest, and Alexander Rush. 2020.
\newblock \href {https://doi.org/10.18653/v1/2020.emnlp-demos.6} {Transformers: State-of-the-art natural language processing}.
\newblock In \emph{Proceedings of the 2020 Conference on Empirical Methods in Natural Language Processing: System Demonstrations}, pages 38--45, Online. Association for Computational Linguistics.

\bibitem[{Ye et~al.(2022{\natexlab{a}})Ye, Gao, Li, Xu, Feng, Wu, Yu, and Kong}]{ye-etal-2022-zerogen}
Jiacheng Ye, Jiahui Gao, Qintong Li, Hang Xu, Jiangtao Feng, Zhiyong Wu, Tao Yu, and Lingpeng Kong. 2022{\natexlab{a}}.
\newblock \href {https://aclanthology.org/2022.emnlp-main.801} {{Z}ero{G}en: Efficient zero-shot learning via dataset generation}.
\newblock In \emph{Proceedings of the 2022 Conference on Empirical Methods in Natural Language Processing}, pages 11653--11669, Abu Dhabi, United Arab Emirates. Association for Computational Linguistics.

\bibitem[{Ye et~al.(2022{\natexlab{b}})Ye, Gao, Wu, Feng, Yu, and Kong}]{ye-etal-2022-progen}
Jiacheng Ye, Jiahui Gao, Zhiyong Wu, Jiangtao Feng, Tao Yu, and Lingpeng Kong. 2022{\natexlab{b}}.
\newblock \href {https://aclanthology.org/2022.findings-emnlp.269} {{P}ro{G}en: Progressive zero-shot dataset generation via in-context feedback}.
\newblock In \emph{Findings of the Association for Computational Linguistics: EMNLP 2022}, pages 3671--3683, Abu Dhabi, United Arab Emirates. Association for Computational Linguistics.

\bibitem[{Zhang et~al.(2022)Zhang, Roller, Goyal, Artetxe, Chen, Chen, Dewan, Diab, Li, Lin, Mihaylov, Ott, Shleifer, Shuster, Simig, Koura, Sridhar, Wang, and Zettlemoyer}]{zhang2022opt}
Susan Zhang, Stephen Roller, Naman Goyal, Mikel Artetxe, Moya Chen, Shuohui Chen, Christopher Dewan, Mona Diab, Xian Li, Xi~Victoria Lin, Todor Mihaylov, Myle Ott, Sam Shleifer, Kurt Shuster, Daniel Simig, Punit~Singh Koura, Anjali Sridhar, Tianlu Wang, and Luke Zettlemoyer. 2022.
\newblock \href {http://arxiv.org/abs/2205.01068} {Opt: Open pre-trained transformer language models}.

\end{thebibliography}
\bibliographystyle{acl_natbib}

\appendix

\section{Appendix}
\label{sec:appendix}

\subsection{Screencast} \label{sec:appendix-screencast}
A screencast about the \methodname{} framework can be found on \href{https://vimeo.com/850491720}{Vimeo}.

\begin{table*}[ht]
\centering
\begin{tabular}{lccccc}
\toprule
\textbf{Dataset} & Data & \multicolumn{4}{c}{\textbf{\# Training examples}} \\
& & 50 & 500 & 1000 & all \\
\midrule
\multirow{2}{*}{TREC-6} & Gold & $42.7 \pm 9.6$ & $93.8 \pm 0.3$ & $95.1 \pm 0.6$ & $97.1 \pm 0.3$ \\
& Generated & $27.5 \pm 11.0$ & $56.2 \pm 3.3$ & $57.9 \pm 1.6$ & $62.6 \pm 3.4$ \\
\bottomrule
\end{tabular}
\caption{Results on TREC-6 with generated questions by GPT-3.5 using 3 few-shot examples (uniformly sampled from 8 possible few-shot examples per class). We observe that the generation performance is worse compared to an equally sized human-annotated dataset. However, the performance increases with the number of examples generated.}
\label{tab:experiment3-dataset_generation_results}
\end{table*}

\begin{table*}[ht]
\centering
\begin{tabular}{lcccccc}
\toprule
\textbf{Dataset} & \textbf{\# few-shot examples} & \multicolumn{5}{c}{\textbf{\# examples per class used in prompt}} \\
& \textbf{per class} & 0 &2 & 3 & 4 & 5 \\
\midrule
\multirow{5}{*}{TREC-6} & 0 & $30.2 \pm 0.6$ & - & - & - & - \\
& 2 & - & $43.0 \pm 3.7$ & - & - & - \\
& 4 & - & $56.0 \pm 0.5$ & $56.3 \pm 2.4$ & $58.3 \pm 2.2$ & - \\
& 8 & - & $52.8 \pm 1.5$ & $58.8 \pm 1.0$ & $58.2 \pm 1.0$ & $64.0 \pm 2.0$ \\
& 16 & - & $58.3 \pm 0.8$ & $59.8 \pm 2.5$ & $58.7 \pm 1.1$ & $54.8 \pm 1.5$ \\
\bottomrule
\end{tabular}
\caption{Results on 500 generated TREC-6 examples with different sizes of few-shot examples and number of few-shot examples included in the prompt. We observe that more few-shot examples result in better performance on the gold annotated test split.}
\label{tab:experiment4-hyperparameter_generation}
\end{table*}

\subsection{Hyperparameters for Experiments} \label{sec:appendix-hyperparameters}

We used AdamW \citep{loshchilov2018decoupled} as our optimizer with a batch size of 16. Further, we used a linear warm-up for 10\% of the optimization steps. We fine-tune \texttt{roberta-base} for question answering with a learning rate of $1e^{-5}$ for two epochs without early stopping. For the \texttt{bert-base-uncased}~PLM, we fine-tune using a learning rate of $2e^{-5}$ for either 5 (if training data has more than 1000 examples), 10 (if training dataset has at least 500 but less than 1001 examples) or 20 epochs (if training data is less than 501 examples). Further, across all experiments, we use 10\% of the data as a validation split for model selection.

\subsection{Generate Label-Conditioned Training Data} \label{sec:appendix-generation_experiment}

This experiment used label-conditioned generation to create new data for the TREC dataset containing six classes. To achieve this, we sampled a small few-shot dataset from the existing training split, consisting of 8 examples per class. During generation, for each label $y$, we included three uniformly sampled few-shot examples associated with that label. We generated 10k data pairs ($\mathbf{x}^g$, $y$) and used them for fine-tuning. It is important to note that the gold-labeled dataset contains only around 3k examples. Thus the column ``all'' refers either to the 10k examples generated with GPT or to the \textasciitilde 3k gold-labeled examples. The experimental setup is identical to \Cref{sec:experiments}.

The results are depicted in \Cref{tab:experiment3-dataset_generation_results}. We observe significant performance drops compared to the re-annotation experiments for TREC from \Cref{sec:experiment1}. For instance, using 10k generated examples achieves a performance level similar to using 50 human-annotated examples (compare to Table~\ref{tab:experiment1-dataset_annotation_results}). However, we note that we performed no prompt optimization techniques or hyperparameter searches in all experiments. Additionally, we generated a uniform distribution of classes, while the gold-labeled dataset is skewed towards certain categories. It is worth mentioning that this class distribution information may not be available in real-world few-shot settings.

\subsection{Impact of Few-Shot Examples on Label-Conditioned Generation} \label{sec:appendix-generation_hyperparameter_experiment}

In this experiment, we generated 500 label-conditioned data pairs for the TREC dataset, following the approach described in \Cref{sec:experiment2}. We conducted a sweeping analysis over two factors: the total number of few-shot examples per class and the number of few-shot examples included in the actual prompt.

The results are depicted in \Cref{tab:experiment4-hyperparameter_generation}. Our findings show that including even a small number of few-shot examples (< 4) yields better results compared to generating without any few-shot examples. Moreover, when we used at least four examples per class, we observed significant improvements in the generation results, from 30.2 to 54.8 in accuracy ($\uparrow 24.6$ pp. in accuracy). Additionally, using more examples in a distinct prompt slightly improved the model performance. We encountered one outlier when using 16 examples per class and including five examples in the prompt for generation, which resulted in lower performance than sampling from 8 few-shot examples per prompt. It is important to note that during this experiment, we did not adjust any hyper-parameters of the LLM for generation, such as temperature or top-k sampling.

\subsection{Instruction-tuning open-source models}\label{sec:appendix-instructiontuning_experiment}

\begin{table}[ht]
\centering
\begin{tabular}{lccr}
\toprule
\textbf{Model} & \textbf{Acc.} & \textbf{(micro) F1} \\
\midrule
LLaMAv2 + Instr. Tuning & 92.4 & 60.0 \\
\midrule
GPT-3.5$^{\ast}$ & 88.4 & 52.5 \\
\bottomrule
\end{tabular}
\caption{Comparison of instruction-tuned LLaMA models with 3-shot GPT-3.5 based on the training split of CoNLL-03. We report accuracy and span-level F1 score the annotation on the validation split. $^{\ast}$: We convert tag sequences to spans in order to prompt the LLM with strings rather than sequence. However, $38\%$ of the validation split annotations have different lengths after tokenization which have been filtered out for a fair comparison.}
\label{tab:experiment5-instruction_tuning}
\end{table}

In this experiment, we compare the annotation performance of OpenAI's GPT-3.5 with an instruction-tuned open-source LLaMA model. To conduct this evaluation, we choose the token classification task on the CoNLL-03 dataset \citep{tjong-kim-sang-de-meulder-2003-introduction}, which generates one label for each token in the input, making it a structured task.

The results are shown in \Cref{tab:experiment5-instruction_tuning}. We observe that using the dataset as-is results in often unusable annotation outputs, primarily due to imprecise formatting. To address this, we convert the token-level labels into spans and prompt the LLM to extract all named entities for the relevant categories. We then transform the found entities into token-level tags by searching for the annotations as substrings of the input text. We compare the performance of this approach with a instruction-tuned LLaMA model on the entire training split of CoNLL-03 by letting both LLMs annotate the validation set.

Unlike the previous evaluation, we did not train and evaluate a smaller PLM on the gold-labeled test set. Instead, we assess the performance between the gold-annotated validation split and the annotations made by the LLM.
Our findings indicate that the annotation quality of instruction-tuned LLMs can significantly improve over OpenAI's GPT, as evident from the higher F1 score. This finding suggests that instruction-tuned models for dataset generation have the potential to facilitate the generation process for complex downstream tasks in future research endeavors.

\end{document}